\documentclass{article}

\usepackage{nips_2017}

\usepackage{amsmath}
\usepackage{amsfonts}
\usepackage{mathtools}
\usepackage{bbm}
\usepackage{amssymb}
\usepackage{esvect}

\usepackage{graphicx}
\usepackage{tikz}
\usepackage{pgfplots}
\usepackage{pgfplotstable}
\usepgfplotslibrary{statistics}
\usepackage{capt-of}
\usepackage{filecontents}

\usepackage[utf8]{inputenc} % allow utf-8 input
\usepackage[T1]{fontenc}    % use 8-bit T1 fonts
\usepackage{hyperref}       % hyperlinks
\usepackage{url}            % simple URL typesetting
\usepackage{booktabs}       % professional-quality tables
\usepackage{amsfonts}       % blackboard math symbols
\usepackage{nicefrac}       % compact symbols for 1/2, etc.
\usepackage{microtype}      % microtypography

\title{Deep Neural Networks for Survival Analysis Based on a Multi-Task Framework}	

\author{
Stephane Fotso\\
Square, Inc. \\
San Francisco, CA, U.S.A.\\
\texttt{stephane@squareup.com} \\
}

\begin{document}
\maketitle

%%%%%%%%%%%%%%%%%%%%%%%%%%%%%%              ABSTRACT            %%%%%%%%%%%%%%%%%%%%%%%%%%%%%%%%%
\begin{abstract}
Survival analysis/time-to-event models are extremely useful as they can help companies predict when a customer will buy a product, churn or default on a loan, and therefore help them improve their ROI. In this paper, we introduce a new method to calculate survival functions using the Multi-Task Logistic Regression (MTLR) model as its base and a deep learning architecture as its core. Based on the Concordance index (C-index) and Brier score, this method outperforms the MTLR in all the experiments disclosed in this paper as well as the Cox Proportional Hazard (CoxPH) model when nonlinear dependencies are found.
\end{abstract}

%\begin{keywords}
%Survival Analysis, Deep Learning, Neural Networks, Keras, TensorFlow
%\end{keywords}

%%%%%%%%%%%%%%%%%%%%%%%%%%%%%%          INTRODUCTION       %%%%%%%%%%%%%%%%%%%%%%%%%%%%%%%%%
\section{Introduction}
Survival analysis is a field in statistics that is used to predict when an event of interest will happen. The field emerged from medical research as a way to model a patient's survival --- hence the term "survival analysis". More recently, it has gained a lot of traction in other industries, such as e-commerce, advertising, telecommunication and financial services, where companies have started using their databases to better understand when their customers will purchase a product \cite{ji2016probabilistic}, churn \cite{perianez2016churn} or default on a loan \cite{Dirick2017}.
 
The most common survival analysis modeling techniques are the Kaplan-Meier (KM) model \cite{kaplan1958nonparametric}  and Cox Proportional Hazard (CoxPH) model \cite{cox1992regression}. The KM model provides a very easy way to compute the survival function of an entire cohort, but doesn't do so for a specific individual. The CoxPH model's approach enables the end-user to predict the survival and hazard functions of an individual based on its feature vector, but exhibits the following limitations: 
\begin{itemize}
\item It assumes that the hazard function, or more precisely the log of the hazard ratio, is powered by a linear combination of features of an individual.
\item It relies on the proportional hazard assumption, which specifies that the hazard function of two individuals has to be constant over time.
\item The exact formula of the model that can handle ties isn't computationally efficient, and is often rewritten using approximations, such as the Efron’s\cite{efron1977efficiency} or Breslow's\cite{breslow1974covariance} approximations,  in order to fit the model in a reasonable time.
\item The fact that the time component of the hazard function (i.e., the baseline function) remains unspecified makes the CoxPH model ill-suited for actual survival function predictions. 
\end{itemize}

To overcome these drawbacks, Chun-Nam Yu et al. \cite{yu2011learning} introduced the MTLR model that can calculate the survival function without any of the aforementioned assumptions or approximations. The one problem that remains is the core of the model is still linear; therefore, it cannot properly model nonlinear dependencies in the dataset. One way to offset the linearity problem is to introduce a deep learning architecture \cite{Goodfellow-et-al-2016}. Jared Katzman et al. \cite{katzman2016deep} and Margaux Luck et al. \cite{luck2017deep} introduced state-of-the-art neural networks within survival analysis, but both still rely heavily on the CoxPH model structure.
 
In this work, we propose a new model: the \textbf{Neural Multi-Task Logistic Regression} (N-MTLR) model. This model relies on the MTLR technique, but its core is powered by a deep learning architecture. Based on the main survival analysis performance metrics, C-index \cite{harrell1982evaluating} and Brier score \cite{graf1999assessment} \cite{gerds2006consistent}, the N-MTLR outperforms the MTLR in all the experiments disclosed in this paper. It also outperforms the CoxPH model when nonlinear dependencies are found in the datasets.
 
In section 2, we give a quick overview of survival analysis, the quantity at stake and a review of the KM and CoxPH models. In section 3, we give details on the MTLR model. In section 4, we introduce the proposed model. The last section looks at the result of the experiments, comparing the CoxPH, MTLR and N-MTLR models by using the C-index and Brier score on artificially generated and real datasets.

%%%%%%%%%%%%%%%%%%%%%   Part 1 - Survival Analysis, an Overview       %%%%%%%%%%%%%%%%%%%%%%%%%%%%
\section{Survival Analysis, an Overview}
Survival analysis corresponds to a collection of statistical modeling methods designed to estimate the time until an event of interest will occur.
The response variable, also called the survival time, is recorded from an initial timestamp until the occurrence of the event of interest or until the subject exits the analysis without experiencing the event; this phenomenon is called \textbf{censoring}.
In case of (right) censoring, we can't know the exact time of the event; the only information at our disposal is that the event will occur after the censoring time.

%%%% 1.1 Notations 
\subsection{Notations}
Let's define the mathematical elements that will be used throughout the rest of this paper:
\begin{itemize}
  \item $T$ is a non-negative random variable, modeling the waiting time until an event occurs. Its probability density function is $f(t)$ and its cumulative density function is $F(t)$. \[F(t) = \int_{-\infty}^t f(s) ds \]
  \item $S(t)$ is the survival function, defined by $S(t) = P[T>t] = 1 - F(t)$, that calculates the probability that the event of interest has not occurred by some time $t$.
  \item $h(t)$ is the hazard function that represents the instantaneous rate of occurrence for the event of interest.\[ h(t) = \lim_{ dt \to 0} \frac{P[t \leq T < t + dt | T \geq t]}{dt}  \text{ or } h(t)= \frac{f(t)}{S(t)} \]
 It is possible to link $S(t)$ and $h(t)$ such that: \[ S(t) = \exp \left(- \int_0^t h(s) ds \right) \] 
 \end{itemize}  

Given a dataset of $N$ samples, we will characterize survival analysis data such that $ \forall i \in [\![1, N]\!]$, $( \vv{x_i}, \delta_i, T_i)$ represents a datapoint.
\begin{itemize}
  \item $  \vv{x_i} \in \mathbbm{R}^p$ is a $p-$dimensional feature vector, defined by $ \vv{x_i} =  [x^i_1,   x^i_2,  \hdots , x^i_p ]$.
  \item $ \delta_i$ is the event indicator such that $\delta_i = 1$, if an event happens and  $\delta_i = 0$ in case of censoring.
  \item $ T_i =  \min(t_i, c_i)$ is the observed time, with $t_i$ the actual event time and $c_i$ the time of censoring. 
 \end{itemize}

%%%% 1.2 Classical Models 
\subsection{Classical Models}
Multiple methods exist to determine $S(t)$ or $h(t)$. This section will summarize the two most popular models.

\subsubsection{Kaplan-Meier Model}
The \textbf{Kaplan-Meier} (KM) estimator is a non-parametric model used to compute the survival function $\hat{S}_{KM}(t)$ of a homogeneous cohort.
Given $N$ units in a cohort, let's assume that there are $J$ distinct actual event times such that $t_1 < t_2 < ... < t_J$ with $J \leq N$, then the survival function $\hat{S}_{KM}(t)$ is given by:
\begin{equation} 
  \tag{Kaplan-Meier}
	\hat{S}_{\text{KM}}(t) = \prod_{t_j \leq t}\left(1- \frac{d_j}{r_j} \right)
\end{equation}
   
where:
\begin{itemize}
  \item $d_j$ is the number of units that experience an event at $t_j$.
  \item $r_j$ is the number of units  at risk (units that haven't experienced the event or haven't been censored yet) in the time interval $[t_{j-1}, t_j)$.
\end{itemize}

\subsubsection{Cox Proportional Hazard Model}
Although easy to compute, the Kaplan-Meier model doesn't take into account feature vectors. Thus, to be able to predict the survival function of a specific unit, it is very common to use the \textbf{Cox Proportional Hazard model} (CoxPH). 
The CoxPH model is a semi-parametric model that focuses on modeling the hazard function $\hat{h}_{\text{CoxPH}}(t, \vv{x_i}) $, by assuming that its time component $\lambda_0(t)$ and  feature component $\displaystyle \eta(\vv{x_i})$ are proportional such that:
\begin{equation} 
  \tag{CoxPH}
	\hat{h}_{\text{CoxPH}}(t, \vv{x_i}) = \lambda_0(t) \eta(\vv{x_i})
\end{equation}

where :
\begin{itemize}
  \item $ \lambda_0(t) $ is the baseline function, which is usually not specified.
  \item $\eta(\vv{x_i})$ is the risk function usually expressed via a linear representation such that \[ \eta(\vv{x_i}) = \exp \left( \sum_{j=1}^p x^i_j\beta_j \right) \text{ with }   \beta_j \text{s being the coefficients to determine.}
  \]
\end{itemize}

%%%%%%%%%%%%%%%%%%%%%   Part 2 - Multi-Task Logistic Regression     %%%%%%%%%%%%%%%%%%%%%%%%%%%%
\section{Multi-Task Logistic Regression (MTLR)}
When it comes to predicting the survival function for a specific unit, the Cox Proportional Hazard Model is usually the go-to model. However, it presents some important drawbacks:
\begin{itemize}
\item It relies on the proportional hazard assumption, which specifies that the hazard function of two individuals has to be constant over time.
\item The exact formula of the model that can handle ties isn't computationally efficient, and is often rewritten using approximations, such as the Efron’s or Breslow's approximations,  in order to fit the model in a reasonable time.
\item The fact that the time component of the hazard function remains unspecified makes the CoxPH model ill-suited for actual survival function predictions. 
\end{itemize}

That's the reason why the \textbf{Multi-Task Logistic Regression} (MTLR) model was developed. It can be seen as a series of logistic regression models built on different time intervals so as to estimate the probability that the event of interest happened within each interval. 

The model can be built using the following steps:

\begin{enumerate}
  \item Dividing the time axis into $J$ time intervals such that $\forall j \in [\![1, J ]\!]$, $ a_j = [ \tau_{j-1}, \tau_j )$ with $ \tau_0 = 0 $ and $\tau_J = \infty$.

\begin{center}
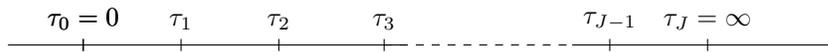

  \begin{tikzpicture}
    \draw (3,0) -- (8.3,0);
    \foreach \x/\xtext in {4/$\tau_0=0$, 5.3/$\tau_1$, 6.6/$\tau_2$, 8/$\tau_3 $}{
      \draw (\x ,-.08) -- (\x,0.08) node[above] {\xtext};
    }
    dashed
        \draw[dashed] (8,0) -- (10.5,0);
    \foreach \x/\xtext in {4/$\tau_0=0$}{
      \draw (\x ,-.08) -- (\x,0.08) node[above] {\xtext};
    }
    
     \draw[->] (10.5,0) -- (14,0);
    \foreach \x/\xtext in {11/$ \tau_{J-1}$, 12.3/$\tau_J = \infty $}{
      \draw (\x ,-.08) -- (\x,0.08) node[above] {\xtext};
    }
  \end{tikzpicture} 
	\captionof{figure}{Subdivisions of the time axis in $J$ intervals}
  \end{center}
  
  \item Building a logistic regression model on each interval $a_j$, with the parameters $\left( \vv{\theta}_j, b_j \right)$   and the response variable $y_j  = 
				 \begin{cases}
				 1 \text{    , if } T \in a_j \text{ i.e., the event happened in the interval } a_j \\
				 0 \text{    , otherwise} \\
    				\end{cases}$. 
But, because we are not analyzing the effects of recurrent events, we need to make sure that when a unit experiences an event on interval $a_s$ with $s \in [\![1, J ]\!]$, its status for the remaining intervals stays the same. Thus, the response vector $\vv{Y}$ is:
    \[ \vv{Y} =  \begin{bmatrix}y_1 = 0 \\  y_2 = 0 \\ \vdots \\ y_{s-1} = 0 \\  y_{s} = 1 \\  \vdots  \\  y_{J} =1 \end{bmatrix} \]
  
Chun-Nam Yu et al. \cite{yu2011learning} proposed the following definitions for the density and survival functions:
 \begin{itemize}
  \item \underline{Density function}:
  \begin{equation}
   \tag{MTLR - density}
  \begin{split}
f(a_s,  \vv{x}) & = P\left[  T \in [\tau_{s-1}, \tau_s) | \vv{x} \right]  \\
	& = \frac{\exp\left(   \sum_{j = s}^{J-1} \vv{x} \cdot \vv{\theta}_j + b_j \right) }{ Z(\vv{x}) } \\
	& = \frac{\exp\left(  (\vv{x} \cdot \textbf{$\Theta$} + \vv{b} )\cdot \textbf{$\Delta$} \cdot \vv{Y}   \right) }{ Z(\vv{x}) } 
\end{split}
  \end{equation}

  \item \underline{Survival function}:
   \begin{equation}
   \tag{MTLR - survival}
	\begin{split}
S(\tau_{s-1}, \vv{x})  = P\left[  T \geq \tau_{s-1}|\vv{x} \right ]  & =  \sum^{J}_{k = s} P\left[  T \in [\tau_{k-1}, \tau_k) | \vv{x} \right ]  \\
 & =  \sum^{J}_{k = s} \frac{ \exp \left(   \sum^{J-1}_{j=s} (\vv{x} \cdot \vv{ \theta}_j + b_j) \right) }{  Z(  \vv{x})}	\\
 & =  \sum^{J}_{k = s} \frac{ \exp \left(  \left(  \vv{x} \cdot \textbf{$\Theta$} + \vv{b} \right) \cdot \textbf{$\Delta$}  \cdot  \vv{Y} \right) }{  Z\left(  \vv{x}\right)}	
\end{split}
  \end{equation}

\end{itemize}

with\footnote{Linear algebra operations simplified in \cite{Jin:15}}:
\begin{itemize}

    \item $ \textbf{$\Theta$}   =  \begin{bmatrix} \theta_{1, 1}  &  \theta_{1, 2}  & \hdots &  \theta_{1, J-1}&  \theta_{1, J} \\  
    												\theta_{2, 1}  &  \theta_{2, 2}  & \hdots &  \theta_{2, J-1}&  \theta_{2, J} \\  
    													    & \vdots    &  \hdots  &\hdots &   &   \\
    												\theta_{p, 1}  &  \theta_{p, 2}   & \hdots &  \theta_{p, J-1}&  \theta_{p, J} \\  
    												 \end{bmatrix} 
    				 =   \left[ \vv{\theta}_1,  \vv{\theta}_2,   \hdots,    \vv{\theta}_{J-1},  \vv{\theta}_J \right]
    				$, a $ \mathbbm{R}^{p \times J}$ coefficients matrix.
    													  
    \item $ \vv{b}  = \begin{bmatrix} b_1, & b_2,   & \hdots &  b_{J-1}, &  b_J  \end{bmatrix} $, a $ \mathbbm{R}^{1 \times J}$ coefficients (bias) matrix.

    \item $  \textbf{$\Delta$}  =  \begin{bmatrix} 1 & 0 & 0 & \hdots & 0 & 0 \\  
    													 					1 & 1  & 0 & \hdots & 0 & 0 \\
    													    					 \vdots  & \vdots  &  \vdots  &\hdots &   \vdots & \vdots   \\
    													 1 & 1  & 1 & \hdots & 0 & 0 \\  
    													 1 & 1  & 1 & \hdots & 1 & 0 \\  
    													  \end{bmatrix} $, a $ \mathbbm{R}^{J \times (J+1)}$ triangular matrix.
    
    \item $ \displaystyle  Z\left(  \vv{x}\right) = \sum_{j=1}^J \exp \left( \sum_{l=j+1}^J \vv{\theta}_l  \cdot \vv{x} + b_l\right)$, a normalization constant.
\end{itemize}
\end{enumerate}

%%%%%%%%%%%%%%%%%%%%%   Part 3 - Neural Multi-Task Logistic Regression     %%%%%%%%%%%%%%%%%%%%%%%%%%%%
\section{Neural Multi-Task Logistic Regression (N-MTLR)}
Although the MTLR model provides similar results as the CoxPH model without having to rely on the assumptions required by the latter, at its core, it is still powered by a linear transformation. Thus, both models fail to capture nonlinear elements from the data and consequently stop yielding satisfactory performances. The use of state-of-the-art deep learning techniques to build survival models that would capture nonlinear relations was introduced by Jared Katzman et al. \cite{katzman2016deep} and Margaux Luck et al. \cite{luck2017deep}, but both approaches still rely on the CoxPH model architecture. The \textbf{Neural Multi-Task Logistic Regression} (N-MTLR) will help solve this issue.

%%%% 3.1 Added value 
\subsection{Added value}
The introduction of the Neural Multi-Task Logistic Regression provide two major improvements: 

\begin{enumerate}
\item The N-MTLR uses a deep learning framework via a multi-layer perceptron (MLP). By replacing the linear core of the MTLR, the N-MTLR brings a lot of flexibility in the modeling, without relying on any CoxPH model assumptions.
\item The model is implemented in Python using the open-source libraries TensorFlow\footnote{TensorFlow: an open-source software library for Machine Intelligence --- \href{https://www.tensorflow.org/}{\url{https://www.tensorflow.org/}}} and Keras\footnote{Keras: the Python Deep Learning library --- \href{https://keras.io/ }{\url{https://keras.io/}}}, which allows the end-user to use many state-of-the-art techniques, such as:
\begin{itemize}
\item Initialization schemes --- i.e., Xavier uniform, Xavier gaussian \cite{pmlr-v9-glorot10a}.
\item Optimization schemes --- i.e., Adam \cite{adam}, RMSprop \cite{Tieleman2012}.
\item Activation functions --- i.e.,  Softplus, ReLU, SeLU.  
\item Miscellaneous operations --- i.e., Batch Normalization \cite{batchn}, Dropout \cite{dropout}. \newline
\end{itemize}
Moreover, to the best of my knowledge, it is the first time that the MTLR is made available in Python. 
\end{enumerate}

%%%% 3.2 Proposed Model
\subsection{Proposed Model}
By using a deep learning approach, the density and survival functions become:

\begin{itemize}
\item  \underline{Density function}: \begin{equation}
\tag{N-MTLR - density}
\begin{split}
f(a_s, \vv{x}) = P\left[  T \in [\tau_{s-1}, \tau_s) | \vv{x} \right ]  & =   \frac{ \exp \left(  \psi( \vv{x})  \cdot \textbf{$\Delta$}  \cdot  \vv{Y} \right) }{ Z\left( \psi( \vv{x})\right) }
\end{split}
\end{equation}

\item  \underline{Survival function}: \begin{equation}
\tag{N-MTLR - survival}
\begin{split}
S(\tau_{s-1}, \vv{x})  =   \sum^{J}_{k = s} \frac{ \exp \left(   \psi( \vv{x})  \cdot  \textbf{$\Delta$}  \cdot  \vv{Y} \right) }{ Z\left( \psi( \vv{x})\right) }	
\end{split}
\end{equation}

\end{itemize}

$   \psi: \mathbbm{R}^p \mapsto \mathbbm{R}^{J} $ is the nonlinear transformation using $\vv{x} \in  \mathbbm{R}^{p}$ feature vector as its input. Its output is a $\mathbbm{R}^{J}$ vector whose values are mapped to the $J$ subdivisions of the time axis, and $  Z\left(  \psi( \vv{x}) \right) = \sum_{j=1}^J \exp \left( \sum_{l=j+1}^J \psi( \vv{x} ) \right) $
\newline

\paragraph{Example} 
Let's consider the example of a feed forward neural network with 2 hidden layers:
\begin{itemize}
  \item Layer $\#$ 1 has  $M_1$ units and $h^{(1)}(x) =\tanh(x)$ as its activation function
  \item Layer $\#$ 2 has $M_2$ units and $h^{(2)}(x) =\text{ReLU}(x)$ as its activation function
\end{itemize}

\tikzset{%
   neuron missing/.style={
    draw=none, 
    scale=3.2,
    text height=0.333cm,
    execute at begin node=\color{black}$\vdots$
  },
}

\begin{center}

\begin{tikzpicture}[x=1.5cm, y=1.5cm, >=stealth]
\node [anchor=west] (note) at (0.6, 1.6) { \text{ $\tanh(x)$}};
\node [anchor=west] (note) at (2.4, 1.6) { $\text{ ReLU}(x)$};

\foreach \m/\l [count=\y] in {1,2,3}
{
 \node [circle,fill=green!50,minimum size=1cm] (input-\m) at (0,2.5-\y) {};
}
\foreach \m/\l [count=\y] in {4}
{
 \node [circle,fill=green!50,minimum size=1cm ] (input-\m) at (0,-2.5) {};
}
 
 \node [neuron missing]  at (0,-1.5) {};

\foreach \m/\l [count=\y] in {1, 2}
{
  \node [circle,fill=red!50,minimum size=1cm ] (hidden1-\m) at (2,2.2-\y) {};
 }
\foreach \m/\l [count=\y] in {3}
  \node [circle,fill=red!50,minimum size=1cm ] (hidden1-\m) at (2,-1.85) {};

 \node [neuron missing]  at (2,-0.8) {};

\foreach \m/\l [count=\y] in {1, 2}
{
  \node [circle,fill=red!50,minimum size=1cm ] (hidden2-\m) at (4,2.2-\y) {};
 } 
\foreach \m [count=\y] in {3}
  \node [circle,fill=red!50,minimum size=1cm ] (hidden2-\m) at (4,-1.85) {};
  
 \node [neuron missing]  at (4,-0.8) {};

\foreach \m [count=\y] in {1}
  \node [circle,fill=blue!50,minimum size=1cm ] (output-\m) at (6,1.5-\y) {};
  
\foreach \m [count=\y] in {2}
  \node [circle,fill=blue!50,minimum size=1cm ] (output-\m) at (6,-0.5-\y) {};

 \node [neuron missing]  at (6,-0.4) {};

\foreach \l [count=\i] in {1,2,3,p}
  \draw [<-] (input-\i) -- ++(-1,0)
    node [above, midway] {$x_{\l}$};

\foreach \l [count=\i] in {1,2}
  \node [below] at (hidden1-\i.north) {$h^{(1)}_{\l}$};
\node [below] at (hidden1-3.north) {$h^{(1)}_{M_1}$};

\foreach \l [count=\i] in {1,2}
  \node [below] at (hidden2-\i.north) {$h^{(2)}_{\l}$};
\node [below] at (hidden2-3.north) {$h^{(2)}_{M_2}$};

\foreach \l [count=\i] in {1,J}
  \draw [->] (output-\i) -- ++(1,0)
    node [above, midway] {$\psi_{ \l}$};

\foreach \i in {1,...,4}
  \foreach \j in {1,...,3}
    \draw [->] (input-\i) -- (hidden1-\j);

\foreach \i in {1,...,3}
  \foreach \j in {1,...,3}
    \draw [->] (hidden1-\i) -- (hidden2-\j);

\foreach \i in {1,...,3}
  \foreach \j in {1,...,2}
    \draw [->] (hidden2-\i) -- (output-\j);

\end{tikzpicture}
\end{center}

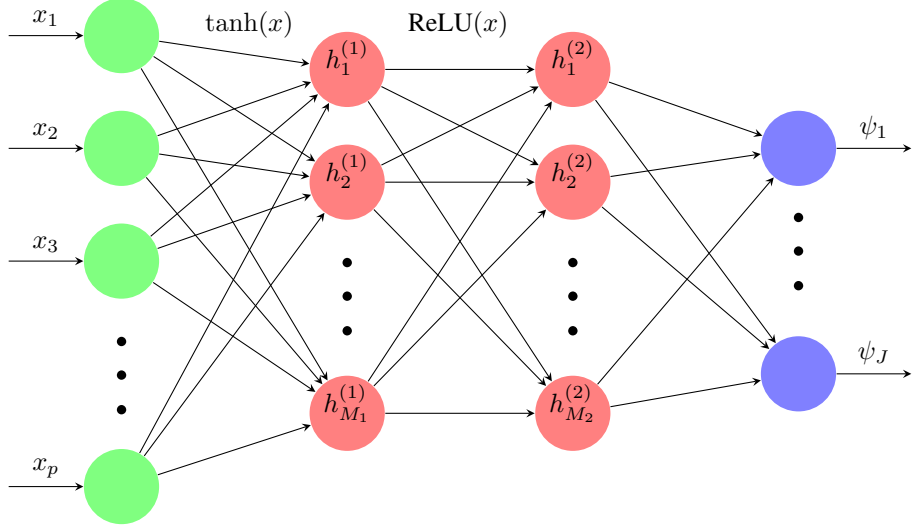
\captionof{figure}{Representation of a 2-hidden layer  transformation }

%%%%%%%%%%%%%%%%%%%%%%%%%%%%%%%     Part 4 - Experiements      %%%%%%%%%%%%%%%%%%%%%%%%%%%%%%%%%%
\section{Experiments}
In this section, we will introduce the main performance metrics in survival analysis and compare the performances of the CoxPH, MTLR and N-MTLR models.

%%%% 4.1 - Performance metrics
\subsection{Performance metrics}
Because right censoring is present in the data, it is necessary to adapt the evaluation metrics that are used in regular machine learning to the framework of survival analysis.

\subsubsection{Concordance Index}
The \textbf{concordance index} or \textbf{C-index} is \textit{a generalization of the area under the ROC curve (AUC)} that can take into account censored data and can be represented by:
		\begin{equation} 
		\tag{C-index}
			\text{C-index} = \frac{ \sum_{i, j} \mathbbm{1}_{T_i > T_j} \cdot \mathbbm{1}_{\eta_i > \eta_j} \cdot \delta_j }{\sum_{i, j} \mathbbm{1}_{T_i > T_j}\cdot \delta_j }
		\end{equation}	  
with:
\begin{itemize}
\item  $\eta_i$, the risk score of a unit $i$
\item  $\mathbbm{1}_{ T_i > T_j } =  \begin{cases} 
      1 & \text{ if } T_i > T_j \\
      0 &  \text{ otherwise }
   \end{cases}
$ and 
$
\mathbbm{1}_{ \eta_i > \eta_j } =  \begin{cases} 
      1 & \text{ if } \eta_i > \eta_j \\
      0 &  \text{ otherwise }
   \end{cases}
$
\end{itemize}

Similarly to the AUC, $\text{C-index}= 1$ corresponds to the best model prediction, and  $\text{C-index} = 0.5$ represents a random prediction.  \newline

\subsubsection{Brier Score}
The \textbf{Brier score} is used to evaluate the accuracy of a predicted survival function at a given time $t$; it represents the \textit{average squared distances between the observed survival status and the predicted survival probability} and is always a number between 0 and 1, with 0 being the best possible value.\newline

Given a dataset of $N$ samples,  $\forall i \in  [\![1, N ]\!],  \left(\vv{x}_i, \delta_i, T_i \right)$ is the format of a datapoint, and the predicted survival function is $ \hat{S}(t, \vv{x}_i), \forall t \in \mathbb{R^+}$:
\begin{itemize}
\item In the absence of right censoring, the Brier score can be calculated such that:
		\begin{equation} 
		\tag{Brier score}
			BS(t) = \frac{1}{N} \sum_{i = 1}^{N} (\mathbbm{1}_{ T_i > t } - \hat{S}(t, \vv{x}_i))^2
		\end{equation}	
 
\item However, if the dataset contains samples that are right censored, then it is necessary to adjust the score by weighting the squared distances using the inverse probability of censoring weights method.
  Let $ \hat{G}(t) = P[C > t ]$ be the estimator of the conditional survival function of the censoring times calculated using the Kaplan-Meier method, where $C$ is the censoring time.
 		\begin{equation}  
		\tag{Weighted Brier score}
			BS(t) =  \frac{1}{N} \sum_{i = 1}^{N} \left( \frac{\left( 0 - \hat{S}(t, \vv{x}_i)\right)^2 \cdot \mathbbm{1}_{T_i \leq t, \delta_i = 1}}{ \hat{G}(T_i^-)} + \frac{ \left( 1 - \hat{S}(t, \vv{x}_i)\right)^2 \cdot \mathbbm{1}_{T_i > t}}{ \hat{G}(t)} \right)
  		\end{equation}	
\end{itemize}

In terms of benchmarks, a useful model will have a Brier score below $0.25$. Indeed,  it is easy to see that if $\forall i \in [\![1, N]\!], \hat{S}(t, \vv{x}_i) = 0.5$, then $BS(t) = 0.25$. \newline

The \textbf{Integrated Brier Score} (IBS) provides an overall calculation of the model performance at all available times.
\begin{equation}  
\tag{IBS}
	\text{IBS} =  \frac{1}{\max(T_i)}   \int_{0}^{\max(T_i)}  BS(t) dt
\end{equation}

%%%% 4.2 - Models comparison
\subsection{Models Comparison}
In this section, we will use the C-index, Brier score and IBS to evaluate the predictive performances of the \textbf{Cox Proportional Hazard model}, the linear \textbf{MTLR} as well as the \textbf{N-MTLR} on the following data:
\begin{enumerate}
\item Simulated survival data
\begin{itemize}
\item Simulated survival data with a linear risk function 
\item Simulated survival data with a nonlinear \textit{square} risk function 
\item Simulated survival data with a nonlinear \textit{gaussian} risk function 
\end{itemize}
\item Real datasets
\begin{itemize}
\item The Worcester Heart Attack Study \cite{hosmer2011applied} (WHAS) dataset\footnote{The dataset can be found at \url{https://github.com/rfcooper/whas/blob/master/whas500.csv}.} 
\item The Veterans’ Administration Lung Cancer \cite{kalbfleisch2011statistical}  (veteran) dataset \footnote{The dataset can be found at \url{https://github.com/cran/survival/raw/master/data/veteran.rda}.} 
\end{itemize}
\end{enumerate}

\subsubsection{Simulated Survival Data}
In this first part, we will generate \cite{generate} a dataset of $3,000$ datapoints. Each observation is represented with $\vv{x} \in \mathbb{R}^3$ such that  $\vv{x} = \begin{bmatrix} x_1, x_2, x_3 \end{bmatrix}$ with: 
\begin{itemize}
\item $x_1 \sim \text{Exp}( \lambda = 0.1 )$
\item $x_2 \sim \text{Normal}( \mu = 10,  \sigma^2 = 5)$ 
\item $x_3 \sim \text{Poisson}(\lambda = 5)$
\end{itemize}
Moreover, we will adjust the censoring time parameters to get around $40\%$ of actual events.

\paragraph{Linear risk function} ---
Each event time $T$ is generated via a \textit{Weibull} distribution such that $\displaystyle T \sim \text{Weibull}(  \lambda = 0.01 \cdot \eta(\vv{x}), p = 2.1) $ with $\displaystyle \eta(\vv{x}) = -0.5 \cdot x_1 + 9 \cdot x_2 + 19 \cdot x_3$ being the risk function. 

For this example, the linear and proportional hazard (PH) assumptions are perfectly valid; therefore, it is not surprising to see that the CoxPH model results are quite good with regard to the C-index and Brier score.  
Moreover, the MTLR and N-MTLR\footnote{Here, the model uses a 1-hidden layer (100 units) with a softmax activation function. It was trained with an Adam optimizer.}  display very similar modeling performances to the CoxPH results, with the N-MTLR slightly outperforming its linear counterpart.

\begin{table}[!htbp]
\centering
\caption{Results for the Weibull survival time dataset with a linear risk function}

\begin{tabular}{lll}
\hline
Model  & C-index (std. error)        & IBS (std. error)    \\ \hline
CoxPH & 0.74 ($1.2 \cdot 10^{-3}$) & 0.10 ($9.8 \cdot 10^{-4}$) \\
MTLR   & 0.74 ($1.2 \cdot 10^{-3}$) & 0.12 ($1.1 \cdot 10^{-3}$) \\
N-MTLR & 0.74 ($1.2 \cdot 10^{-3}$) & 0.10 ($7.9 \cdot 10^{-4}$) \\ \hline
\end{tabular}
\end{table}

%% Brier Score - Results for Linear Weibull %%

\begin{center}
 \begin{tikzpicture}[scale=0.85, transform shape]
 \centering
  \begin{axis}
    [
    title= Models comparison using C-index (median value),
    xtick={1,2,3},
    xticklabels={CoxPH, MTLR, N-MTLR},
    boxplot/draw direction=y,
    ytick={0.5, 0.6, 0.7, 0.8, 0.9, 1.0},
    yticklabels={0.5, 0.6, 0.7, 0.8, 0.9, 1.0},
    ymin=0.5,ymax=1,
    ]
    \addplot+[
    boxplot prepared={
 box extend=0.6,
median = 0.744,
upper quartile = 0.755,
lower quartile = 0.733,
upper whisker = 0.782,
lower whisker = 0.702,
    },      color=blue,
    ] coordinates {};
\node at (axis cs:1,  0.74) [anchor=north east, above,outer sep=10pt ] { (0.74) };

    \addplot+[ boxplot prepared={
 box extend=0.6,
median = 0.742,
upper quartile = 0.752,
lower quartile = 0.725,
upper whisker = 0.787,
lower whisker = 0.692,
    } , color=red
    ] coordinates {};
\node at (axis cs:2,  0.74) [anchor=north east, above,outer sep=10pt ] { (0.74) };

    \addplot+[
    boxplot prepared={
 box extend=0.6,
median = 0.743,
upper quartile = 0.756,
lower quartile = 0.730,
upper whisker = 0.787,
lower whisker = 0.695,
    },       color=black ,
    ] coordinates {};
\node at (axis cs:3,  0.74) [anchor=north east, above,outer sep=10pt ] { (0.74) };

  \end{axis}
  
\end{tikzpicture}
\begin{tikzpicture}[scale=0.85, transform shape]
\centering
    \begin{axis}[
    		legend pos= south east,
    		title= Models comparison using Brier score,
    ytick={0, 0.1,  0.2, 0.3},
    		]
    
    \addplot [blue, dotted, thick, mark=otimes*]  table [x=time, y=coxph] {linear.dat};
    \addlegendentry{CoxPH}
    
    \addplot [red, dashed, mark=x]  table [x=time, y=mtlr] {linear.dat};
    \addlegendentry{Linear MTLR}
    
    \addplot [black, mark=triangle*]  table [x=time, y=nmtlr]{linear.dat};
    \addlegendentry{Neural MTLR}
    
    \end{axis}

\end{tikzpicture}
	
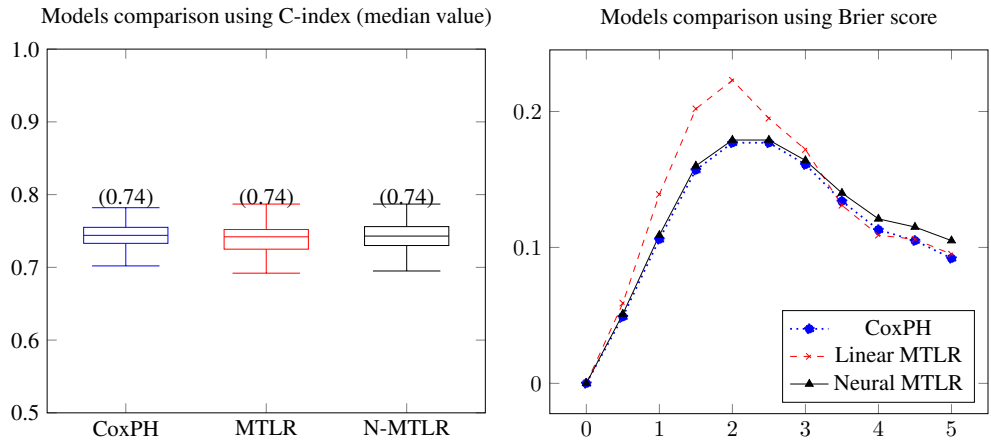
\captionof{figure}{C-index and Brier score --- Weibull survival times with linear risk function}
  \end{center}

\paragraph{Nonlinear \textit{square} risk function} ---
Here the event time $T$ is generated via a \textit{Weibull} distribution such that $\displaystyle T \sim \text{Weibull}(  \lambda = 0.1 \cdot \eta(\vv{x}), p = 2.1) $ with $\displaystyle \eta(\vv{x}) = \left(-0.5 \cdot x_1 + 9 \cdot x_2 + 19 \cdot x_3\right)^2$. 

In this example, it is easy to see that the linear and PH assumptions do not hold anymore. Consequently, the CoxPH model displays less-than-satisfactory results. Moreover, as the core of the MTLR model is linear, it will show similar results as the CoxPH model. However, thanks to the neural networks embedded in the N-MTLR\footnote{Here, the model uses a 2-hidden layer MLP: Layer $\#1 $: 100 units and a softmax activation function; Layer $\#2$:  100 units and a ReLU activation function. It was trained with an Adamax optimizer.}, the model manages to capture the nonlinear dependencies in the data and outperform its linear peers.

\begin{table}[!htbp]
\centering
\caption{Results for the Weibull survival time dataset with a square risk function}
\begin{tabular}{lll}
\hline
Model  & C-Index (std. error)       & IBS (std. error)   \\ \hline
CoxPH & 0.53 ($1.8 \cdot 10^{-3}$) & 0.14 ($6.3 \cdot 10^{-4}$) \\
MTLR   & 0.53 ($1.8 \cdot 10^{-3}$) & 0.18 ($9.3 \cdot 10^{-4}$) \\
\textbf{N-MTLR} & \textbf{0.70 ($1.9 \cdot 10^{-3}$)} & \textbf{0.12 ($5.2 \cdot 10^{-4}$) }\\ \hline
\end{tabular}
\end{table}

%% Brier Score - Results for Square Weibull %%

\begin{center}
 \begin{tikzpicture}[scale=0.85, transform shape]
 \centering
  \begin{axis}
    [
    title= Models comparison using C-index (median value),
    xtick={1,2,3},
    xticklabels={CoxPH, MTLR, N-MTLR},
    boxplot/draw direction=y,
    ytick={0.5, 0.6, 0.7, 0.8, 0.9, 1.0},
    yticklabels={0.5, 0.6, 0.7, 0.8, 0.9, 1.0},
    ymin=0.5,ymax=1,
    ]
    \addplot+[
    boxplot prepared={
 box extend=0.6,
median = 0.531,
upper quartile = 0.547,
lower quartile = 0.515,
upper whisker = 0.593,
lower whisker = 0.501,
    },      color=blue,
    ] coordinates {};
\node at (axis cs:1,  0.56) [anchor=north east, above,outer sep=10pt ] { (0.53) };

    \addplot+[ boxplot prepared={
 box extend=0.6,
median = 0.533,
upper quartile = 0.550,
lower quartile = 0.520,
upper whisker = 0.590,
lower whisker = 0.500,
    } , color=red
    ] coordinates {};
\node at (axis cs:2,  0.56) [anchor=north east, above,outer sep=10pt ] { (0.53) };

    \addplot+[
    boxplot prepared={
 box extend=0.6,
median = 0.696,
upper quartile = 0.711,
lower quartile = 0.679,
upper whisker = 0.737,
lower whisker = 0.641,
    },       color=black ,
    ] coordinates {};
\node at (axis cs:3,  0.70) [anchor=north east, above,outer sep=10pt ] { (0.70) };

  \end{axis}
  
\end{tikzpicture}
\begin{tikzpicture}[scale=0.85, transform shape]
\centering
    \begin{axis}[
    		legend pos= north east,
    		title= Models comparison using Brier score,
    		ytick={0, 0.1,  0.2, 0.3},
    		]
    
    \addplot [blue, dotted, thick, mark=otimes*]  table [x=time, y=coxph] {square.dat};
    \addlegendentry{CoxPH}
    
    \addplot [red, dashed, mark=x]  table [x=time, y=mtlr] {square.dat};
    \addlegendentry{Linear MTLR}
    
    \addplot [black, mark=triangle*]  table [x=time, y=nmtlr]{square.dat};
    \addlegendentry{Neural MTLR}
    
    \end{axis}

\end{tikzpicture}
	
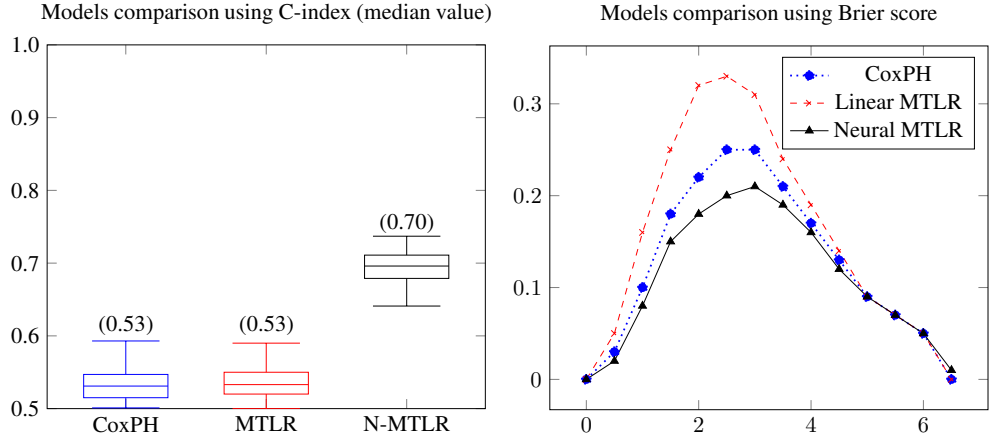
\captionof{figure}{C-index and Brier score --- Weibull survival times with a square risk function}
  \end{center}

\paragraph{Nonlinear \textit{gaussian} risk function} ---
Here, the event time $T$ is generated via a \textit{Weibull} distribution such that $T \sim \text{Weibull}(  \lambda = 0.1 \cdot \eta(\vv{x}), p = 2.1) $ with $ \eta(\vv{x}) = \exp \left( -\frac{\left(-0.5 \cdot x_1 + 9 \cdot x_2 + 19 \cdot x_3\right)^2}{2} \right)$.  Similarly to the previous example, the N-MTLR\footnote{Here, the model uses a 2-hidden layer MLP: Layer $\#1$:  128 units and a ReLU activation function; Layer $\#2$ : 128 units and a softmax activation function. It was trained with a RMSprop optimizer.} outperforms the two other models.

\begin{table}[!htbp]
\centering
\caption{Results for the Weibull survival time dataset with a gaussian risk function}
\begin{tabular}{lll}
\hline
Model  & C-Index (std. error)       & IBS (std. error)   \\ \hline
CoxPH & 0.52 ($1.2 \cdot 10^{-3}$) & \textbf{0.13 ($1.2 \cdot 10^{-3}$)} \\
MTLR   & 0.52 ($1.2 \cdot 10^{-3}$) & 0.16 ($1.5 \cdot 10^{-3}$) \\
\textbf{N-MTLR }& \textbf{0.60 ($2.2\cdot 10^{-3}$)}  & \textbf{0.13 ($1.1 \cdot 10^{-3}$) }\\ \hline
\end{tabular}
\end{table}

%% Brier Score - Results for Gaussian Weibull %%

\begin{center}
 \begin{tikzpicture}[scale=0.85, transform shape]
 \centering
  \begin{axis}
    [
    title= Models comparison using C-index (median value),
    xtick={1,2,3},
    xticklabels={CoxPH, MTLR, N-MTLR},
    boxplot/draw direction=y,
    ytick={0.5, 0.6, 0.7, 0.8, 0.9, 1.0},
    yticklabels={0.5, 0.6, 0.7, 0.8, 0.9, 1.0},
    ymin=0.5,ymax=1,
    ]
    \addplot+[
    boxplot prepared={
 box extend=0.6,
median = 0.513,
upper quartile = 0.522,
lower quartile = 0.505,
upper whisker = 0.541,
lower whisker = 0.500,
    },      color=blue,
    ] coordinates {};
\node at (axis cs:1,  0.52) [anchor=north east, above,outer sep=10pt ] { (0.51) };

    \addplot+[ boxplot prepared={
 box extend=0.6,
median = 0.518,
upper quartile = 0.527,
lower quartile = 0.507,
upper whisker = 0.555,
lower whisker = 0.500,
    } , color=red
    ] coordinates {};
\node at (axis cs:2,  0.53) [anchor=north east, above,outer sep=10pt ] { (0.52) };

    \addplot+[
    boxplot prepared={
 box extend=0.6,
median = 0.595,
upper quartile = 0.611,
lower quartile = 0.582,
upper whisker = 0.649,
lower whisker = 0.542,
    },       color=black ,
    ] coordinates {};
\node at (axis cs:3,  0.63) [anchor=north east, above,outer sep=10pt ] { (0.60) };

  \end{axis}
  
\end{tikzpicture}
\begin{tikzpicture}[scale=0.85, transform shape]
\centering
    \begin{axis}[
    		legend pos= north east,
    		title= Models comparison using Brier score,
    		ytick={0, 0.1,  0.2, 0.3},
    		]
    
    \addplot [blue, dotted, thick, mark=otimes*]  table [x=time, y=coxph] {gaussian.dat};
    \addlegendentry{CoxPH}
    
    \addplot [red, dashed, mark=x]  table [x=time, y=mtlr] {gaussian.dat};
    \addlegendentry{Linear MTLR}
    
    \addplot [black, mark=triangle*]  table [x=time, y=nmtlr]{gaussian.dat};
    \addlegendentry{Neural MTLR}
    
    \end{axis}

\end{tikzpicture}
	
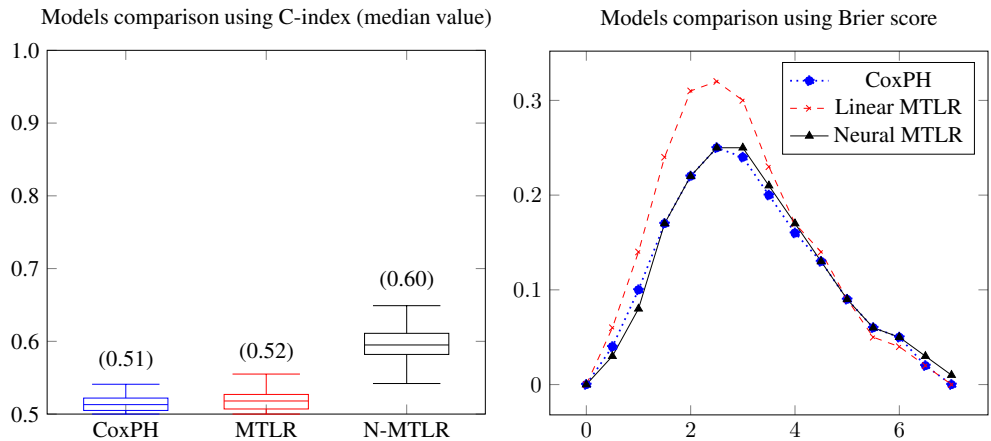
\captionof{figure}{C-index and Brier score --- Weibull survival times with a gaussian risk function}
  \end{center}

\subsubsection{Real datasets}
We will now compare the three different models on real survival datasets.

\paragraph{Worcester Heart Attack Study (WHAS)} ---
The Worcester Heart Attack Study dataset displays the survival times of patients after they've had a heart attack. As the dataset doesn't seem to contain any obvious nonlinear dependencies, the three models perform quite similarly. We note that the N-MTLR\footnote{Here, the model uses a 2-hidden layer MLP: Layer $\#1$:  100 units and a softmax activation function; Layer $\#2$ : 30 units and a ReLU activation function. It was trained with a RMSprop optimizer.} slightly outperforms the MTLR and CoxPH models.

  \begin{table}[!htbp]
\centering
\caption{Results for the WHAS dataset}
\begin{tabular}{lll}
\hline
Model  & C-Index (std. error)  &  IBS (std. error)   \\ \hline
CoxPH &0.79 ($3.2 \cdot 10^{-3}$) & 0.17 ($1.9 \cdot 10^{-3}$) \\
MTLR   & 0.78 ($3.4 \cdot 10^{-3}$) & 0.17 ($2.4 \cdot 10^{-3}$) \\
\textbf{N-MTLR} & \textbf{0.80} ($3.6\cdot 10^{-3}$)  & 0.17 ($1.9 \cdot 10^{-3}$) \\ \hline
\end{tabular}
\end{table}

%% Brier Score - Results for WHAS %%

\begin{center}
 \begin{tikzpicture}[scale=0.85, transform shape]
 \centering
  \begin{axis}
    [
    title= Models comparison using C-index (median value),
    xtick={1,2,3},
    xticklabels={CoxPH, MTLR, N-MTLR},
    boxplot/draw direction=y,
    ytick={0.5, 0.6, 0.7, 0.8, 0.9, 1.0},
    yticklabels={0.5, 0.6, 0.7, 0.8, 0.9, 1.0},
    ymin=0.5,ymax=1,
    ]
    \addplot+[
    boxplot prepared={
 box extend=0.6,
median = 0.790,
upper quartile = 0.812,
lower quartile = 0.768,
upper whisker = 0.862,
lower whisker = 0.713,
    },      color=blue,
    ] coordinates {};
\node at (axis cs:1,  0.82) [anchor=north east, above,outer sep=10pt ] { (0.79) };

    \addplot+[ boxplot prepared={
 box extend=0.6,
median = 0.776,
upper quartile = 0.796,
lower quartile = 0.750,
upper whisker = 0.854,
lower whisker = 0.687,
    } , color=red
    ] coordinates {};
\node at (axis cs:2,  0.81) [anchor=north east, above,outer sep=10pt ] { (0.78) };

    \addplot+[
    boxplot prepared={
 box extend=0.6,
median = 0.795,
upper quartile = 0.817,
lower quartile = 0.772,
upper whisker = 0.869,
lower whisker = 0.712,
    },       color=black ,
    ] coordinates {};
\node at (axis cs:3,  0.83) [anchor=north east, above,outer sep=10pt ] { (0.80) };

  \end{axis}
  
\end{tikzpicture}
\begin{tikzpicture}[scale=0.85, transform shape]
\centering
    \begin{axis}[
    		legend pos= south east,
    		title= Models comparison using Brier score,
    		ytick={0, 0.1,  0.2, 0.3},
    		]
    
    \addplot [blue, dotted, thick, mark=otimes*]  table [x=time, y=coxph] {whas.dat};
    \addlegendentry{CoxPH}
    
    \addplot [red, dashed, mark=x]  table [x=time, y=mtlr] {whas.dat};
    \addlegendentry{Linear MTLR}
    
    \addplot [black, mark=triangle*]  table [x=time, y=nmtlr]{whas.dat};
    \addlegendentry{Neural MTLR}
    
    \end{axis}

\end{tikzpicture}
	
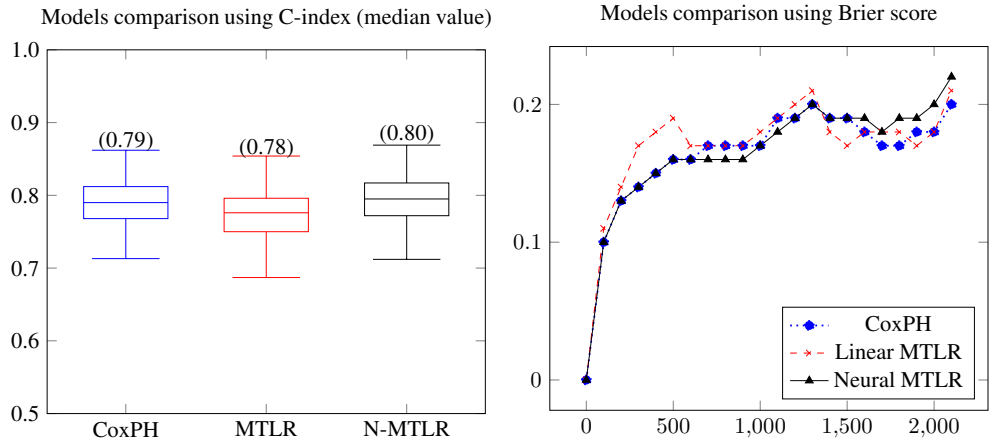
\captionof{figure}{C-index and Brier score --- WHAS}
  \end{center}

\paragraph{Veterans’ Administration Lung Cancer (veteran)} ---
The Veterans’ Administration Lung Cancer dataset displays the survival time of male patients with advanced inoperable lung cancer. The study was conducted by the US Veterans Administration.
Once again, this dataset doesn't seem to contain any obvious nonlinear dependencies; therefore, the three models\footnote{Here, the model uses a 2-hidden layer MLP: Layer $\#1$:  50 units and a softmax activation function; Layer $\#2$ : 10 units and a tanh activation function. It was trained with a RMSprop optimizer.}  perform similarly. 

\begin{table}[!htbp]
\centering
\caption{Results for the veteran dataset}
\begin{tabular}{lll}
\hline
Model  & C-Index (std. error)       & IBS (std. error)   \\ \hline
CoxPH       & 0.78 ($3.1 \cdot 10^{-3}$) & 0.07 ($1.5 \cdot 10^{-4}$) \\
MTLR        & 0.78 ($3.0 \cdot 10^{-3}$) & 0.07 ($1.6 \cdot 10^{-4}$) \\
N-MTLR    & 0.78 ($3.6 \cdot 10^{-3}$) & 0.07 ($1.5 \cdot 10^{-4}$) \\ \hline
\end{tabular}
\end{table}

%% Brier Score - Results for veteran %%

\begin{center}
 \begin{tikzpicture}[scale=0.85, transform shape]
 \centering
  \begin{axis}
    [
    title= Models comparison using C-index (median value),
    xtick={1,2,3},
    xticklabels={CoxPH, MTLR, N-MTLR},
    boxplot/draw direction=y,
    ytick={0.5, 0.6, 0.7, 0.8, 0.9, 1.0},
    yticklabels={0.5, 0.6, 0.7, 0.8, 0.9, 1.0},
    ymin=0.5,ymax=1,
    ]
    \addplot+[
    boxplot prepared={
 box extend=0.6,
median = 0.790,
upper quartile = 0.817,
lower quartile = 0.761,
upper whisker = 0.877,
lower whisker = 0.680,
    },      color=blue,
    ] coordinates {};
\node at (axis cs:1,  0.85) [anchor=north east, above,outer sep=10pt ] { (0.78) };

    \addplot+[ boxplot prepared={
 box extend=0.6,
median = 0.781,
upper quartile = 0.803,
lower quartile = 0.757,
upper whisker = 0.865,
lower whisker = 0.704,
    } , color=red
    ] coordinates {};
\node at (axis cs:2,  0.85) [anchor=north east, above,outer sep=10pt ] { (0.78) };

    \addplot+[
    boxplot prepared={
 box extend=0.6,
median = 0.786,
upper quartile = 0.812,
lower quartile = 0.758,
upper whisker = 0.872,
lower whisker = 0.679,
    },       color=black ,
    ] coordinates {};
\node at (axis cs:3,  0.85) [anchor=north east, above,outer sep=10pt ] { (0.78) };

  \end{axis}
  
\end{tikzpicture}
\begin{tikzpicture}[scale=0.85, transform shape]
\centering
    \begin{axis}[
    		legend pos= north east,
    		title= Models comparison using Brier score,
    		ytick={0, 0.02, 0.04,  0.06, 0.08, 0.10, 0.12, 0.14, 0.16},
    		yticklabels={0, 0.02, 0.04,  0.06, 0.08, 0.10, 0.12, 0.14, 0.16},
     ymin=0,ymax=0.17,
   		]
    
    \addplot [blue, dotted, thick, mark=otimes*]  table [x=time, y=coxph] {veteran.dat};
    \addlegendentry{CoxPH}
    
    \addplot [red, dashed, mark=x]  table [x=time, y=mtlr] {veteran.dat};
    \addlegendentry{Linear MTLR}
    
    \addplot [black, mark=triangle*]  table [x=time, y=nmtlr]{veteran.dat};
    \addlegendentry{Neural MTLR}
    
    \end{axis}

\end{tikzpicture}
	
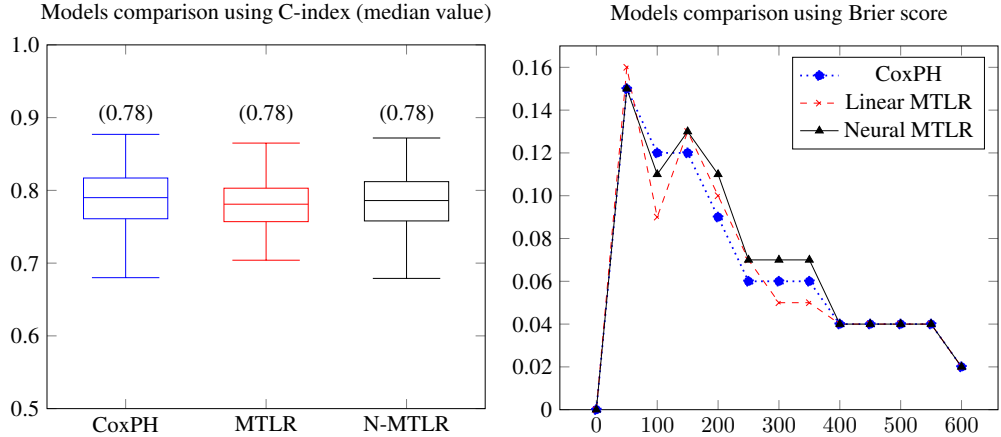
\captionof{figure}{C-index and Brier score --- veteran}
  \end{center}

 %%%%%%%%%%%%%%%%%%%%%%%%%%%%%%          CONCLUSION       %%%%%%%%%%%%%%%%%%%%%%%%%%%%%%%%%
\section{Conclusion}
In conclusion, this paper demonstrates that the Neural Multi-Task Logistic Regression consistently outperforms the MTLR and yields similar or better results than the Cox Proportional Hazard model. Moreover, when nonlinear dependencies can be found in the data, the N-MTLR displays a clear advantage over the other methods studied in this work. 
Finally, because it is written in Python using the packages Keras and TensorFlow, it takes advantage of the state-of-the-art modeling techniques used in deep learning.

\section*{Acknowledgments}
I would  like to thank Berkeley Almand-Hunter, Emily McLinden, Eric Jeske, Jackie Brosamer, Jochen Bekmann, Philip Spanoudes and Thomson Nguyen of the Square Capital Data Science team for their help with the project, as well as Jessie Fetterling and Sara Vera.

%%%%%%%%%%%%%%%%%%%%%%%%%%%%%%          REFERENCES       %%%%%%%%%%%%%%%%%%%%%%%%%%%%%%%%%
\bibliographystyle{plain}

%%%%%%%%%%%%%%%%%%%%%%%%%%%%%%      END DOCUMENT    %%%%%%%%%%%%%%%%%%%%%%%%%%%%%%%%%
\end{document}